\documentclass{article}

\usepackage{PRIMEarxiv}

\usepackage[utf8]{inputenc} % allow utf-8 input
\usepackage[T1]{fontenc}    % use 8-bit T1 fonts
\usepackage{hyperref}       % hyperlinks
\usepackage{url}            % simple URL typesetting
\usepackage{booktabs}       % professional-quality tables
\usepackage{amsfonts}       % blackboard math symbols
\usepackage{nicefrac}       % compact symbols for 1/2, etc.
\usepackage{microtype}      % microtypography
\usepackage{lipsum}
\usepackage{fancyhdr}       % header
\usepackage{graphicx}       % graphics
\graphicspath{{media/}}     % organize your images and other figures under media/ folder

\usepackage{amsmath}
\usepackage{amssymb}
\usepackage{amsfonts}
\usepackage{amsthm}
\usepackage{subfig}
\usepackage{longtable}
\usepackage{multirow}
\usepackage{appendix}
\usepackage{booktabs}

\title{SG-DeepONet: Source-generalized deep operator learning for full waveform inversion
}

\author{
  Zekai Guo \\
  Nanjing University of \\
  Aeronautics and Astronautics \\
  Nanjing, China\\
  \texttt{guozekai@nuaa.edu.cn} \\
   \And
  Lihui Chai \\
   Sun Yat-sen University \\
   Guangzhou, China\\
  \texttt{chailihui@mail.sysu.edu.cn} \\
  \And
  Ye Li \\
  Nanjing University of \\
  Aeronautics and Astronautics \\
  Nanjing, China\\
  \texttt{yeli20@nuaa.edu.cn} \\
}

\begin{document}
\maketitle

\begin{abstract}
Full waveform inversion (FWI) aims to reconstruct subsurface velocity models from observed seismic wavefields and has recently benefited from advances in deep learning (DL).
The performance of DL-based FWI critically depends on the diversity of training data, yet existing datasets such as OpenFWI rely on fixed or weakly varying source conditions, limiting their ability to represent realistic seismic scenarios and hindering source generalization.
To address this issue, we construct a new source-variable seismic dataset, termed SVFWI, by systematically varying the frequencies and horizontal locations of multiple surface sources.
SVFWI is further divided into three subsets that respectively model frequency variations, location variations, and their combined effects, providing a challenging benchmark in data-driven FWI.
We further propose SG-DeepONet, a novel DeepONet-based encoder–decoder framework tailored for FWI.
The branch network extracts multi-scale time–frequency features from seismic observations, the trunk network explicitly embeds source physical parameters, and an interactive decoding network enables effective nonlinear fusion and high-fidelity velocity reconstruction.
Extensive experiments on SVFWI demonstrate that SG-DeepONet achieves superior inversion accuracy and robustness under varying source conditions compared with existing DL-based FWI methods.
\end{abstract}

% keywords can be removed

\section{Introduction}

\begin{figure}
  \centering
  \includegraphics[width=0.47\textwidth]{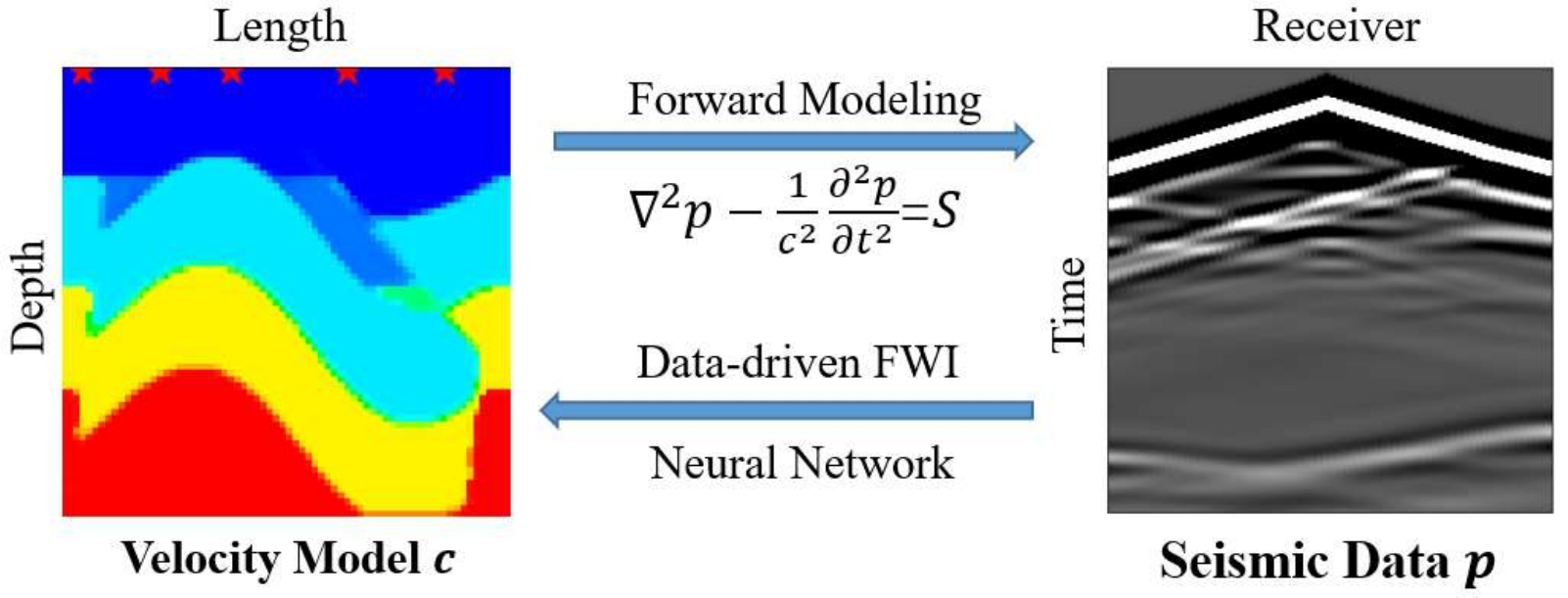} 
  \caption{Forward modeling and data-driven FWI. Red stars represent five seismic sources placed on the ground surface to generate waves. Velocity model (representing the subsurface image) indicates the seismic wave speeds in the subsurface medium. Seismic data is obtained from receivers placed on the ground surface.} 
  \label{img:forward-FWI} 
\end{figure}

% \IEEEPARstart{C}{omputational} wave imaging (CWI) extracts hidden structure and physical properties of a volume of material by analyzing wave signals passing through it~\cite{lin2024physics}.
In seismic wave inversion, we use full waveform inversion (FWI) to reconstruct the subsurface images (velocity models) based on the observed seismic wave signals~\cite{virieux2009overview}.
%Full waveform inversion (FWI) plays a crucial role in understanding subsurface velocity structures\cite{virieux2009overview}.
Not only does FWI have applications in oil and gas exploration, but also find utility in 
%earthquake early warning and 
medical ultrasound imaging domain~\cite{guasch2020full}.
In FWI, forward modeling involves numerically solving acoustic wave equations using an assumed velocity model to simulate the spatial and temporal distribution of seismic waves.
This process relies on well-established physical principles.
Conversely, the seismic inverse problem focuses on reconstructing the unknown subsurface images from observed seismic data.
Fig.~\ref{img:forward-FWI} illustrates these two processes.
% For seismic inversion which is the inversion problem, we need to utilize observed surface measurements to infer the subsurface structure.
Traditional FWI methods, such as iterative optimization algorithms, are computationally expensive due to the repeated evaluations of forward modeling.
%or the high dimensions of wave equations.
Moreover, the nonlinear nature of the least-squares loss function gives rise to a multitude of local minima, thereby complicating optimization, especially when the accuracy of the initial model is insufficient~\cite{zhang2020fwi}.

\begin{figure}
   \centering
     \subfloat[]{\includegraphics[width=0.15\textwidth]{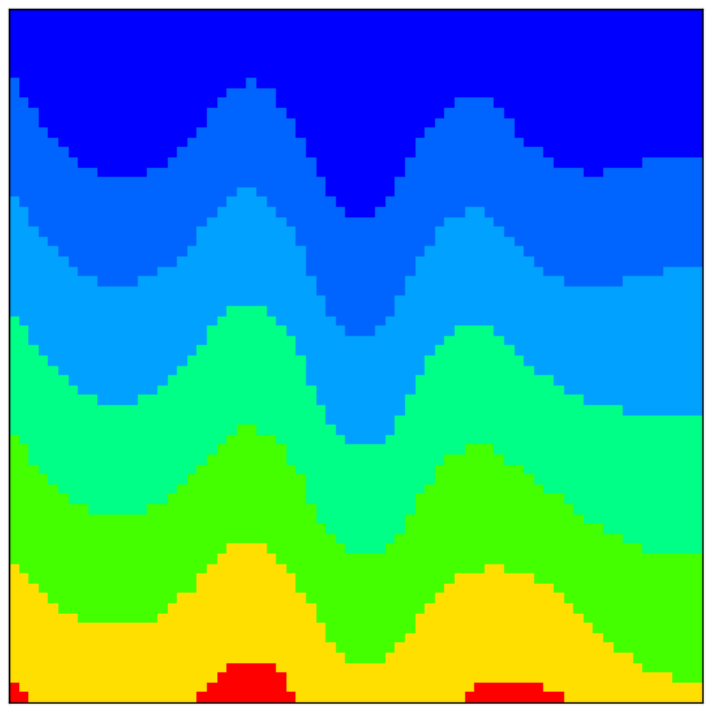}}\hspace{1mm}
     \subfloat[]{\includegraphics[width=0.15\textwidth]{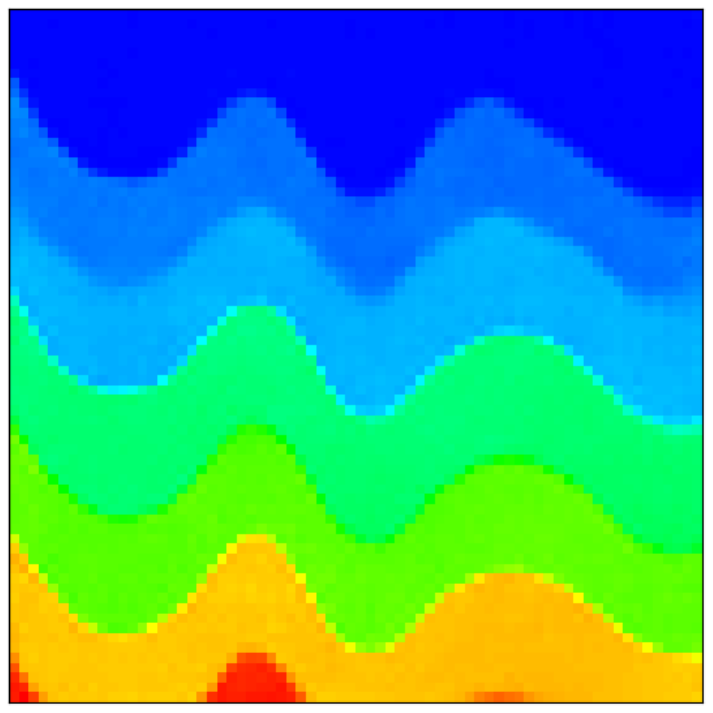}}
     \hspace{1mm}
     \subfloat[]{\includegraphics[width=0.15\textwidth]{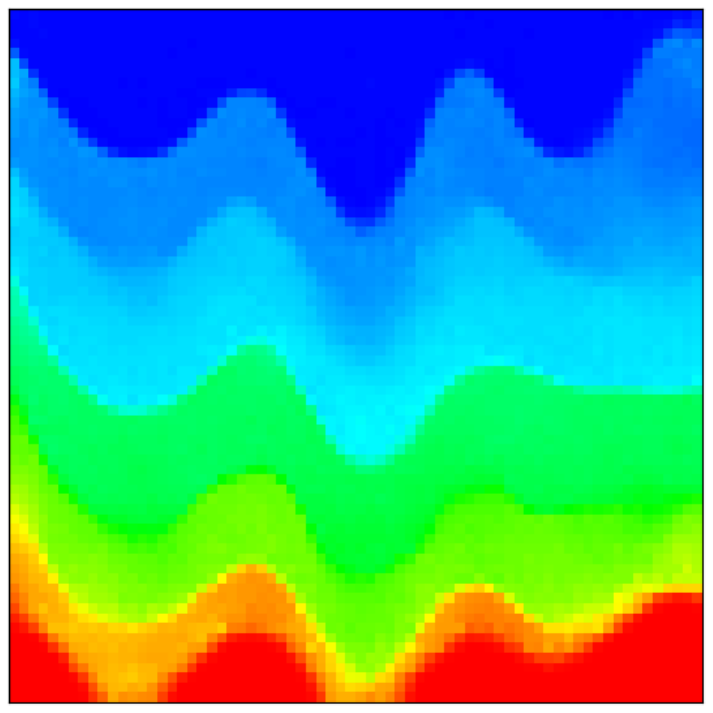}}
     \caption{
     % (a) Ground Truth. The predicted velocity models of InversionNet tested on seismic data where the frequencies of five sources are: (b) all 15 Hz. (c) 14, 15, 16, 15, and 14 Hz, respectively.
     Illustration of the sensitivity of InversionNet to source frequency variations.
(a) Ground Truth.
(b) Predicted result using seismic data with fixed source frequencies (15 Hz).
(c) Predicted result using seismic data with slightly perturbed source frequencies (14, 15, 16, 15, and 14 Hz).
     %(We use the pre-trained InversionNet model from the original paper\cite{wu2019inversionnet}.)}
     }
     %(We use the InversionNet model trained on OpenFWI which has fixed source frequency at 15 Hz.)  }
     \label{fig:intro-generalization}
\end{figure}

With the development of computational power and artificial intelligence, deep learning (DL) approaches demonstrate immense potential in FWI.
DL-FWI provides an end-to-end workflow without the need for manual intervention, featuring strong nonlinear fitting capabilities that can effectively learn the mapping from seismic observations to velocity models.
There are several popular network architectures, such as
convolutional neural networks (CNNs) and generative adversarial networks (GANs), which have shown the significant superiority in FWI~\cite{zhang2019velocitygan, wu2019inversionnet, yang2023fwigan}.
Despite their success, existing DL-based FWI methods exhibit a critical limitation: their predictions are highly sensitive to variations in seismic source conditions.
In practical applications, even slight changes in source frequency or location can lead to substantial degradation in inversion accuracy.

As illustrated in Fig.~\ref{fig:intro-generalization}, models trained under fixed source settings often fail to generalize when applied to unseen source configurations, revealing a fundamental gap between current DL-FWI approaches and realistic seismic acquisition scenarios.
This limitation is closely related to the design of existing public datasets.
OpenFWI~\cite{deng2022openfwi}, the first large-scale open dataset for DL-based FWI, has significantly accelerated progress in this field.
Subsequently, \cite{zhu2023fourier} extended OpenFWI by introducing datasets with variable source frequencies and locations.
However, although source parameters vary across the dataset, all sources within a single seismic record still share the same frequency, which does not reflect realistic multi-source acquisition settings where different sources may operate at distinct frequency bands simultaneously.
In practice, combining low- and high-frequency sources is essential for comprehensive subsurface characterization:
low frequencies constrain large-scale background structures, while high frequencies capture fine-scale geological details.

Motivated by this observation, we develop SVFWI, a source-variant full waveform inversion dataset built upon OpenFWI. SVFWI simultaneously varies source frequencies and horizontal locations among multiple sources, explicitly capturing source diversity within individual observations.
This source-generalization dataset exposes a key challenge for existing DL models, such as InversionNet~\cite{wu2019inversionnet}, whose performance deteriorates significantly under such source variations.

To address this challenge, we propose a novel deep operator learning framework termed SG-DeepONet.
The time–frequency feature encoding branch network acts as a data-driven encoder, leveraging WTConv and convolutional layers to extract multi-scale time–frequency representations from raw seismic waveforms. In parallel, the source-parameterized trunk network performs a nonlinear embedding of source parameters, producing a high-dimensional conditional representation that characterizes source-dependent physical information.
The two latent representations are fused through element-wise interaction and decoded by an interactive branch–trunk decoding network, which maps the coupled features back to the velocity model domain via a transpose-convolution decoder, enabling effective reconstruction while preserving both observational and source-related information.

\section{Related work}

The DL-FWI approach has attracted considerable attention in the geophysical and machine learning communities due to its potential to reconstruct high-resolution velocity models.
 % Unlike conventional FWI methods that rely on wave-equation-based modeling, DL-FWI leverages the nonlinear representation capabilities of deep neural networks, making it a highly nontrivial and ill-posed problem. 
 To enhance the reliability and accuracy of the predicted velocity models, incorporating physical priors or domain knowledge into the deep learning framework has proven to be beneficial. Depending on the degree of such physical knowledge integration, DL-FWI approaches can be broadly categorized into three groups.

\paragraph{Pure Physics-Driven Methods.}
For traditional methods, we infer the velocity model based on gradient-based optimization methods and governing physics and equations.
They require a reasonably accurate initial velocity model to ensure convergence. If the initial guess deviates significantly from the true velocity distribution, the optimization process is likely to get trapped in local minima, leading to inaccurate results.
\cite{datta2016estimating} use the sparse velocity model obtained from very fast simulated annealing as the starting model, which avoids getting stuck in local minimum to the extent possible.
\cite{zhang2012wave} reduce the artifacts in the gradients through a new wave-energy-based precondition method, providing a faster convergence speed.
\cite{lin2014acoustic} propose a modified total-variation regularization which obtains improving accuracy compared with conventional regularization term.
Moreover, conventional FWI is typically designed to optimize a single velocity model at a time. As a result, any change in the subsurface structure necessitates a complete rerun of the computationally intensive inversion process.
Physics-driven methods necessitate repeated forward modeling and iterative refinement of the estimated velocity model, making them both time-consuming and resource-intensive.

\paragraph{Pure Data-Driven Methods.}
% With the development of DL, it has garnered significant research interest in FWI. 
%\cite{jin2018learn, wang2018velocity, sun2021deep} show the potential ability of CNN applying in seismic inversion.
In data-driven approaches, the inversion process is carried out entirely through learned neural networks, without relying explicitly on wave-equation-based physical modeling.
The potential of CNNs in seismic inversion is demonstrated in studies~\cite{jin2018learn,wang2018velocity,sun2021deep}.
InversionNet and InversionNet3D with encoder-decoder structure have an excellent performance~\cite{wu2019inversionnet,zeng2021inversionnet3d}.
These methods typically encode the seismic data into high-dimensional implicit feature representations, from which the subsurface velocity model is subsequently reconstructed.
DD-Net~\cite{zhang2024dd}, based on U-Net~\cite{ronneberger2015u}, uses two decoder to generate velocity model and layer boundaries, respectively.
The first decoder aims to generate the velocity model and maintain a small error compared to the ground truth.
The second decoder focuses on the accuracy of the layer boundaries in the velocity model, and generate a two-channel image to simulates the contour details.
ABA-FWI introduces three dedicated components that guide the network to focus more attention on the boundaries within the velocity model~\cite{xu2024aba}.
\cite{li2023high} introduce an attention convolutional-neural-network-based velocity inversion algorithm.
%~\cite{zhang2020djoint} develop an adjoint-driven deep-learning FWI, which is not sensitive to the cycle-skipping problem.
%The encoder makes the input 2D-seismic data flattened and then decoder reconstructs the velocity model from the flattened data.
%\cite{zhang2019velocitygan,wang2023deep, yao2023regularization} adopt the concept of generative adversarial network in FWI.
% The concept of generative adversarial network  are also adopted to FWI ~\cite{zhang2019velocitygan,wang2023deep,yao2023regularization}.
VelocityGAN, an indirect training control method, adopts the concept of GANs \cite{zhang2019velocitygan}.
The generator is used to predict velocity model, and the discriminator distinguishes between predicted and real outcomes to increase accuracy.
\cite{yang2023well} take low-resolution velocity models, migration images, and well-log velocities as inputs to build high-resolution velocity models.
%The first decoder aims to generate the velocity model and maintain a small error compared to the ground truth.
%The second decoder focuses on the accuracy of the layer boundaries in the velocity model, and generate a two-channel image to simulates the contour details.
%Then they use a joint loss function which consists of the mse and cross-entropy loss, correspond to the outputs of the two encoders respectively.
% The aforementioned methods have shown the efficiency in seismic inversion, while it has obvious limitation in generalization ability.
Neural operator networks have also achieved success in seismic inversion. 
They employ neural networks to learn an implicit operator that maps seismic data to velocity models, corresponding to the acoustic wave equation which governs the propagation of seismic waves.
% ~\cite{zhu2023fourier,li2023solving,haghighat2024deeponet}.
~\cite{li2023solving} propose a paralleled FNO (PFNO) for FWI.
En-DeepONet is a variant of DeepONet, which adds summation and subtraction instead of only single dot-product operation~\cite{haghighat2024deeponet}. 
This architecture overcomes the problem that the solution has spatial shifts corresponding to the movement of earthquake the source location.
Fourier-DeepONet~\cite{zhu2023fourier} utilizes a more complicated structure as the decoder at the end of the vanilla DeepONet, which consists of FNO and U-FNO.

\paragraph{Hybrid Data-Driven and Physics-Driven Methods.}
These methods do not impose such stringent requirement on data due to the physical prior knowledge \cite{lin2023physics}. 
UPFWI \cite{jin2021unsupervised} proposes a unsupervised deep learning method, integrating forward modeling and CNN in a loop.
%~\cite{rasht2022physics} present an algorithm applying physics-informed neural networks (PINNs) \cite{raissi2019physics} to acoustic wave equation and encoding the prior knowledge of subsurface in neural network. 
\cite{zhu2022integrating} represent the velocity model through a generative neural network and then takes it into PDEs solvers.
\cite{wang2023prior} develop a new paradigm for FWI regularized by generative diffusion models.
FWIGAN integrates the wave equation with a discriminative network ~\cite{yang2023fwigan}.
\cite{rasht2022physics,zhang2023seismic, mardan2024physics} introduce prior knowledge by applying physics-informed neural networks.
% \cite{rasht2022physics} introduces PINNs for FWI. 
In these approach, two neural networks are employed to separately approximate the pressure field $p$ and the velocity model $v$. 
The acoustic wave equation is then incorporated as a soft constraint into the loss function, allowing the entire framework to be trained in an unsupervised manner.
%~\cite{mardan2024physics} introduce PINNs with an attention mechanism.
However, further research is needed to address challenges such as slow inversion speed, low accuracy and poor model generalization.

\begin{table*}[h]
\caption{
Detailed description of the proposed SVFWI dataset, including its three subsets: SVFWI-F, SVFWI-L, and SVFWI-FL.
%whereas the frequencies are uniform in the dataset presented by ~\cite{zhu2023fourier}.}
}
%The obvious distinction is that the five sources simultaneously have different frequencies varying from 5 to 25 Hz in our datasets, while the frequencies are the same in \cite{zhu2023fourier}. }
\label{table:datasets}
\setlength\tabcolsep{0.4cm} 
\centering
% {\rmfamily
\begin{tabular}{l|lrrrrr} 
\toprule
%\multicolumn{2}{c}{Dataset} & Source A & Source B & Source C & Source D & Source E \\\hline
\multirow{2}*{Dataset} & $f$ (Hz) & \multirow{2}*{Source A} & \multirow{2}*{Source B} & \multirow{2}*{Source C} & \multirow{2}*{Source D} & \multirow{2}*{Source E} \\
\cline{2-2}
~ & $xs$ (m) & ~ & ~ & ~ & ~ & ~  \\\hline
\multirow{2}*{OpenFWI} & frequency & 15 & 15 & 15 & 15 & 15 \\
 ~ & location & 0 & 172.5 & 345 & 517.5 & 690 \\\hline
\multirow{2}*{SVFWI-F}  & frequency & [5, 25] & [5, 25] & [5, 25] & [5, 25] & [5, 25] \\
 ~ & location & 0 & 172.5 & 345 & 517.5 & 690 \\\hline
\multirow{2}*{SVFWI-L}  & frequency & 15 & 15 & 15 & 15 & 15 \\
 ~ & location & [0, 50] & [122.5, 222.5] & [295, 395] & [467.5, 567.5] & [640, 690] \\\hline
\multirow{2}*{SVFWI-FL} & frequency & [5, 25] & [5, 25] & [5, 25] & [5, 25] & [5, 25] \\
 ~ & location & [0, 50] & [122.5, 222.5] & [295, 395] & [467.5, 567.5] & [640, 690] \\
\bottomrule
\end{tabular}
% }

\end{table*}

\section{Preliminaries}

\subsection{Full waveform inversion}
The acoustic wave equation plays a crucial role in FWI as it describes the propagation of sound waves in subsurface media.
In this work, we focus on 2-D acoustic wave equation with uniform density, and the second-order PDE as follows:
\begin{equation}\label{acoustic wave equation}
\nabla^2 p(x, z, t) - \frac{1}{c(x, z)^2} \frac{\partial^2 p(x, z, t)}{\partial t^2} = S_{xs,f}(x, z, t),
\end{equation}
where $t$ is time, $p$ is pressure wavefield and $c$ is velocity model.
The $f$ and $xs$ are the frequency and horizontal location of seismic source $S$.
Subsequently, $x$ denotes the horizontal location and $z$ denotes the depth in the subsurface. 
In this work, we make seismic sources on horizontal surface, so the vertical location is 0.
Receivers are also typically placed on the surface, which are used to accept seismic data.
And pressure wavefield $p$ with $z=0$ represent seismic data.

We employ a seismic forward modeling algorithm, according to Eq.~\eqref{acoustic wave equation}, to compute the pressure wavefield from the given velocity model. 
This algorithm is implemented using a finite-difference scheme with zero initial conditions. 
In addition, absorbing boundary conditions are applied to account for wave attenuation and to suppress artificial reflections at the edges of the simulation domain.
Forward modeling means to calculate pressure wavefield $p$ from velocity model $c$ , and this expression is defined as: $p=g(c)$.
In contrast, FWI aims to learn the inversion mapping:
%, we have no direct mathematical equation to describe: 
\begin{equation}\label{eq:inversion}
c=g^{-1}(p).
\end{equation}

\begin{figure*}
  \centering
  \includegraphics[width=0.95\textwidth]{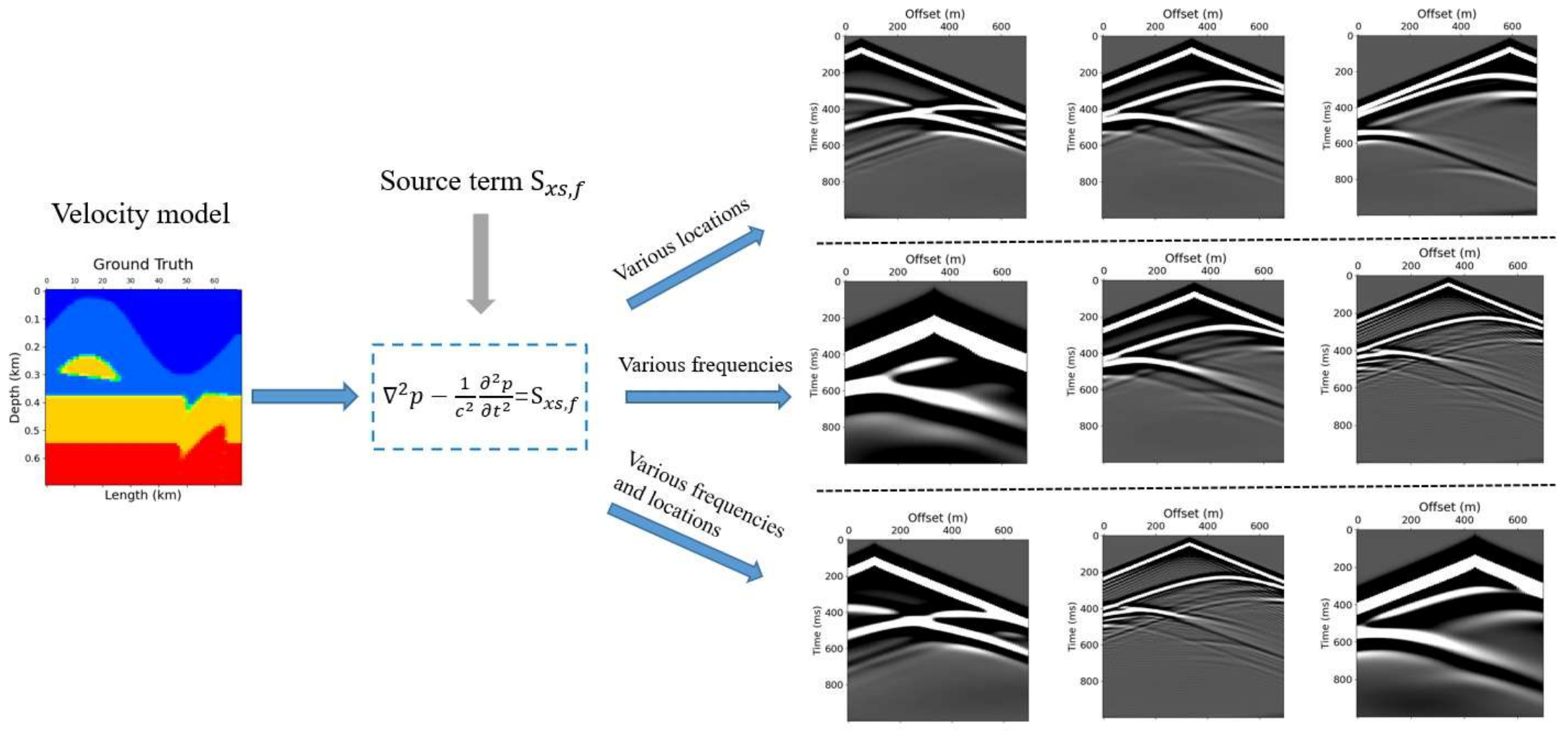} 
  \caption{The generation methods for the SVFWI-L, SVFWI-F, and SVFWI-FL datasets. Row 1 (SVFWI-L): Varying the horizontal locations of the seismic sources while keeping frequency fixed. Row 2 (SVFWI-F): Varying the source frequencies while keeping the source locations fixed. Row 3 (SVFWI-FL): Simultaneously varying both source locations and frequencies.} 
  \label{img:dataset} 
\end{figure*}

\subsection{Source-variant full waveform inversion dataset}\label{datasets}
OpenFWI is a widely used open dataset for data-driven FWI, providing large-scale seismic observations with diverse subsurface velocity models \cite{deng2022openfwi}.
However, in OpenFWI, the source configurations are fixed, with predefined horizontal locations (0, 172.5, 345, 517.5, and 690 m) and a single dominant frequency of 15 Hz, resulting in limited variability in the source term $S_{xs,f}$ in Eq.~\eqref{acoustic wave equation}.
To alleviate this limitation, subsequent datasets introduce variability in source locations and frequencies. Nevertheless, an important practical scenario remains underexplored: multiple seismic sources deployed on the surface may exhibit distinct frequency contents within a single observation, rather than sharing a uniform frequency.

To explicitly model source diversity, we construct a new seismic dataset, termed SVFWI (Source-Variable Full Waveform Inversion), in which the five surface sources are allowed to vary in frequency and/or horizontal location. Velocity models from multiple geological families in OpenFWI are employed, and the corresponding seismic records are generated via forward modeling.
This process can be described as:
\begin{equation}\label{eq:inversion2}
[c, S_{xs,f}] \mapsto p(x, z=0, t),
\end{equation}
where the pressure wavefield $p$ recorded at the surface $(z=0)$ serves as the seismic observation, the velocity model 
$c$ is treated as the ground-truth label, and the source parameters $S_{xs,f}$ are incorporated as explicit physical priors.

The SVFWI dataset is further divided into three subsets for systematic source generalization evaluation: SVFWI-F (varying frequencies), SVFWI-L (varying source locations), and SVFWI-FL (simultaneous variation of frequencies and locations).
This dataset design provides a more challenging and realistic benchmark for data-driven FWI methods under diverse source conditions.

Table~\ref{table:datasets} comprehensively presents detailed information about these datasets.
Fig.~\ref{img:dataset} illustrates the seismic observation signals generated from the same velocity model using sources with different frequencies and horizontal locations. The variations in source parameters lead to diverse wavefield responses, highlighting the sensitivity of seismic data to both source configuration and subsurface structure.
The main objective of our experiment is the identification and reconstruction of subsurface interfaces and faults.
Consequently, we select four datasets for this purpose: FlatVel-B, CurveVel-A, FlatFault-B, and CurveFault-A.
Based on the complexity of the subsurface structure, these datasets are divided into two versions: easy (A) and difficult (B).

\textbf{FlatVel and CurveVel} are designed to detect interfaces which describe the contours of subsurface structures and define the velocity properties of rock layers. 
These datasets provide velocity models that include both flat and curved layers with clear boundaries. 
In version A, the velocity values within the layers gradually increase with depth, while in version B, they are randomly distributed.

\textbf{FlatFault and  CurveFault} are developed for fault identification which is essential for identifying, characterizing, and locating reservoirs.
Faults caused by the shifted rock layers can trap fluid hydrocarbons and form reservoirs. These datasets include discontinuities in the velocity maps caused by the faults.
Version B exhibits more discontinuities and severe velocity changes than version A.

\section{Methodology}
%In this section, we present a novel DeepONet-based network with encoder-decoder (named Inversion-DeepONet).
%We use DeepONet architecture to learn the mapping from seismic data $p$ to velocity model $c$ due to its generalization ability 
%and success in solving inverse problems.
%The encoder-decoder is used to effectively construct velocity model.
We propose SG-DeepONet, a novel DeepONet-based architecture that integrates an encoder–decoder structure and explicitly incorporates physical source parameters into the network design.
We leverage the DeepONet framework to learn the mapping from seismic data $p$ to velocity model $c$, due to its robust generalization ability and proven success in addressing inverse problems.

\begin{figure}
  \centering
  \includegraphics[width=0.47\textwidth]{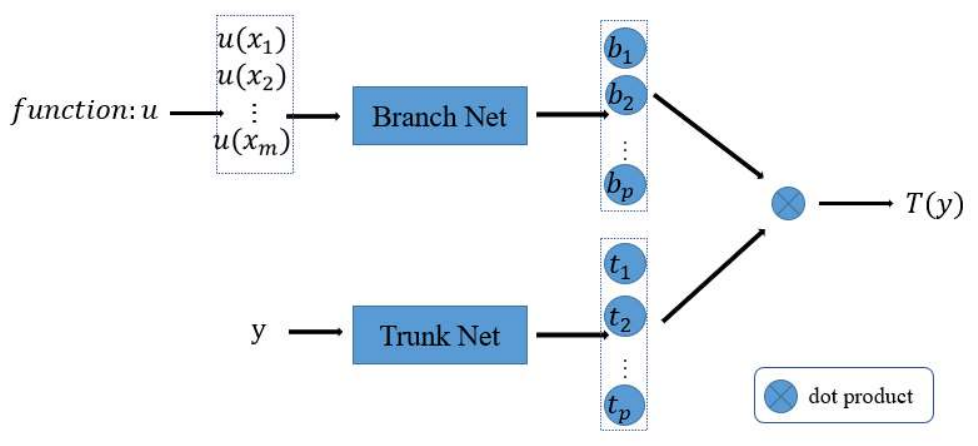} 
  \caption{Structure of vanilla DeepONet. In the context of seismic waveform inversion, the observed seismic data 
$p$ can be treated as the input function $v$ to the branch network, while the velocity model $c$ serves as the output function $T$. The spatial coordinates $(x,z)$ in $c(x,z)$ are used as inputs to the trunk network.} 
  \label{img:deeponet} 
\end{figure}

\begin{figure*}
  \centering
  \includegraphics[width=0.95\textwidth]{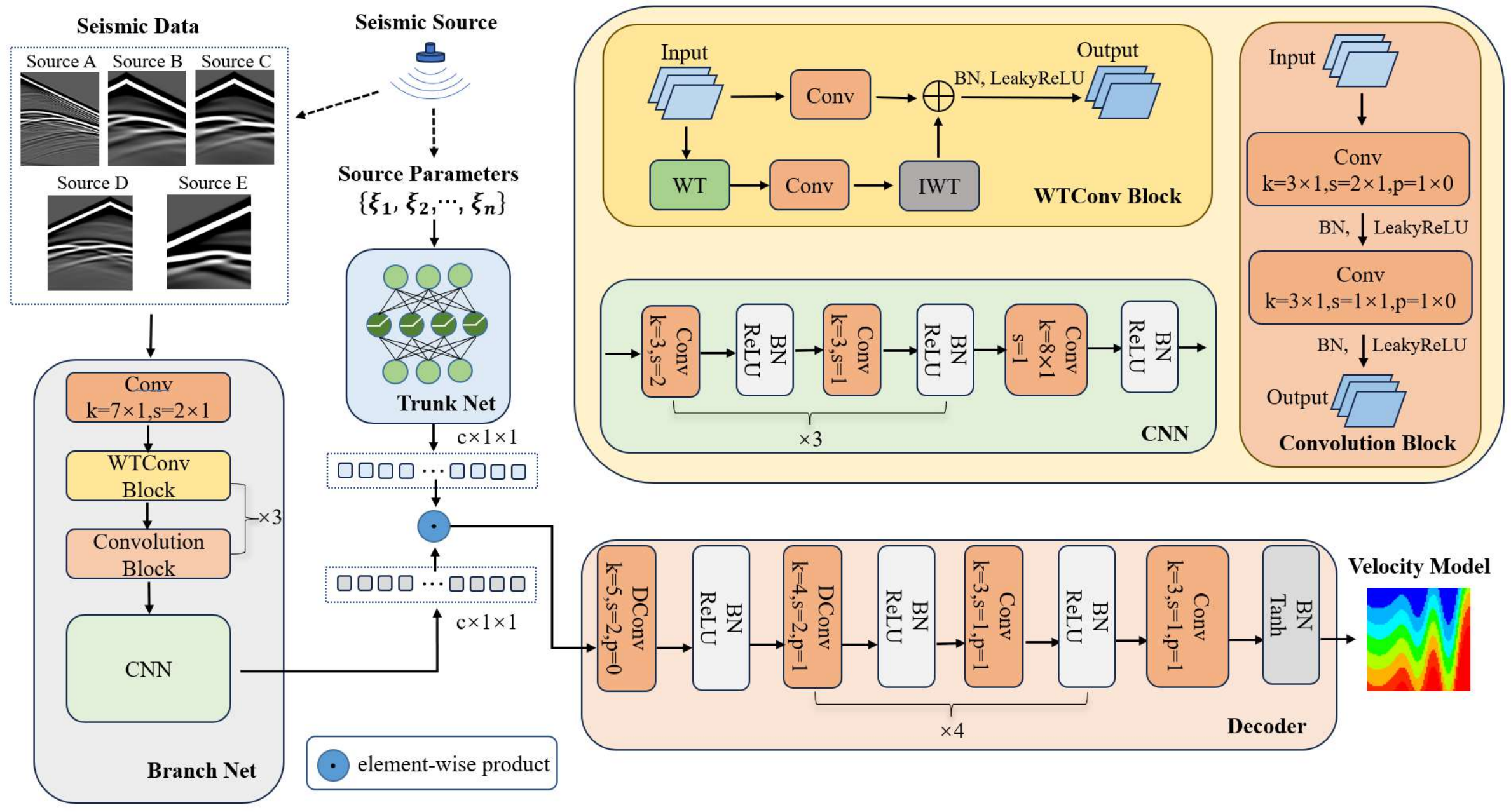} 
  \caption{Architecture of SG-DeepONet for learning the inversion mapping from seismic data to velocity model. Within the WTConv block, a WTConv layer with a decomposition depth of one is adopted.} 
  \label{img:architecture} 
\end{figure*}

\subsection{Vanilla DeepONet}
In recent years, operator learning method %based on universal approximation theorem 
offers a superior performance in solving inversion problems \cite{lu2021learning,li2020fourier}. 
Seismic wave inversion can be considered as operator learning from seismic observational data $p$  to velocity models $c$ according to Eq. \eqref{acoustic wave equation}.
DeepONet is developed to learn operators between infinite-dimensional function spaces, 
which consists of two subnetworks: branch net and trunk net.
The branch net utilizes a set of discrete function values $\left\{v(x_1), v(x_2),..., v(x_m)\right\}$ as inputs and the trunk net takes the location $y$ of final output function $T(y)$ as inputs.
The operator $G$ learned by the neural network can be represented as:
\begin{equation}
   G : v ( x ) \rightarrow T ( y ). 
\end{equation}

Fig. \ref{img:deeponet} shows the architecture of vanilla DeepONet and the network’s output can be presented as:
\begin{equation}
T(v)(y) = \sum_{i=1}^p b_i(v(x_1), v(x_2),..., v(x_m)) \cdot t_i(y), 
\end{equation}
where $\left\{b_1, b_2,..., b_p\right\}$ and $\left\{t_1, t_2,..., t_p\right\}$ are p-dimensional outputs of two subnetworks respectively.
% In recent years, this operator learning method based on universal approximation theorem, 
% offers a superior performance in solving inversion problems \citep{lu2021learning,li2020fourier}. 
However, there are three limitations of vanilla DeepONet applied in FWI:

\textbf{a) Lack of explicit modeling of frequency-dependent seismic features.}
Seismic signals are inherently multi-frequency and strongly non-stationary, with different frequency components corresponding to distinct geological structures. Low-frequency components are crucial for reconstructing large-scale and deep velocity structures, while high-frequency components primarily encode interfaces, faults, and sharp velocity contrasts.
However, vanilla DeepONet relies on standard convolutional or fully connected layers that operate mainly in the time or spatial domain, without mechanisms to explicitly decompose or represent seismic signals across different frequency bands. 
% As a result, frequency-dependent characteristics are learned only implicitly, making it difficult for the network to disentangle the contributions of different frequency components. 
This limitation hinders the simultaneous recovery of deep background structures and fine-scale geological details, ultimately degrading inversion robustness and interpretability.

\textbf{b) Insufficient utilization of source-related physical information.}
% According to the definition of Equation~\eqref{acoustic wave equation}, source term exerts a profound influence on seismic inversion processes.
% In practical applications, the source can be selected and manipulated by humans, offering a degree of control.
% However, the physical features (e.g. location and frequency) of sources have not been optimally leveraged when we employ the vanilla DeepONet directly. 
According to the definition of Eq.~\ref{acoustic wave equation}, source term exerts a profound influence on seismic inversion processes.
In practical applications, the source can be selected and manipulated by humans, offering a degree of control.
However, the physical features of sources (e.g., location and frequency) have not been optimally leveraged in existing approaches. 
For instance, both vanilla DeepONet~\cite{lu2021learning} and InversionNet~\cite{wu2019inversionnet} neglect these source-related priors.
As a result, when the source frequency or location varies, the predicted velocity models change significantly, indicating poor robustness and generalization.

\textbf{c) Limitations of Linear Branch–Trunk Fusion.}
Utilizing the dot product as the final output will give rise to the problem of suboptimal performance \cite{mei2024fully}.
% According to our experimental results, using dot product yields  unsatisfactory results, particularly at the layer boundaries within velocity model. 
(1) From an experimental perspective, vanilla DeepONet's use of a dot product for decoding often leads to suboptimal outputs in seismic inversion, especially where sharp velocity contrasts occur at geological layer boundaries. 
(2) From a theoretical perspective, the suboptimality of vanilla DeepONet largely stems from its use of a simple dot product at the final layer to merge the latent representations of the input function 
$u$ and the spatial coordinate $y$. 
This design limits the interaction between 
$u$ and $y$ to just a single step at the output, without deeper or progressive fusion across layers, making it difficult for the network to capture complex cross-dependencies.

% \begin{figure*}[h]
%    \centering
%     \includegraphics[width=0.9\textwidth]{FIG/FF-F.png}
%      \caption{The predictions of three models tested on FlatFault-B from FWI-F. 
%      For this example, the five source locations are 0, 172.5, 345.0, 517.5.2, and 690.0 m (red stars in ground truth). The five source frequencies are 17.3, 10.9, 15.4, 5.3, and 5.5 Hz.}
%      \label{fig:FWI-F}
% \end{figure*}

%However, taking the inner product as final output will give rise to the problem of suboptimal performance \cite{mei2024fully}.
%In addition, the coordinates of velocity model fed to trunk net are the same for every velocity model which is not beneficial for model training. 
%Hence, vanilla DeepONet generates unsatisfactory results, especially at the layer boundaries according to our experimental results. 

\subsection{Architecture of SG-DeepONet}\label{heuristic}
The overall architecture of the proposed SG-DeepONet is illustrated in Fig.~\ref{img:architecture}.
It adopts an encoder–decoder framework and consists of three key components:a time–frequency feature encoding branch network, a source-parameterized trunk network, and an interactive branch–trunk decoding network.

\subsubsection{Time–frequency feature encoding branch network}
\begin{figure}
  \centering
  \includegraphics[width=0.47\textwidth]{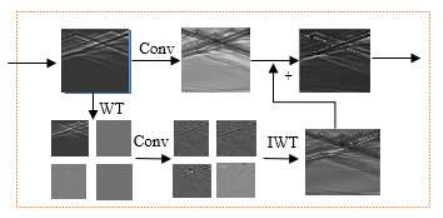} 
  \caption{Schematic illustration of the WTConv layer, highlighting the wavelet-based multi-scale decomposition and convolutional processing of frequency-dependent features.} 
  \label{img:WTconv} 
\end{figure}

In the branch component of SG-DeepONet, we design a wavelet-convolution-based encoding network to extract multi-scale time–frequency features from seismic observations.
Conventional multilayer perceptron (MLP) and CNNs mainly operate in the time or spatial domain and rely on local weighted aggregation. While such architectures are effective at capturing localized patterns, they lack explicit mechanisms to model frequency-dependent characteristics of seismic signals.
To address this issue, we introduce wavelet convolution (WTConv) \cite{finder2024wavelet} into the branch network, enabling explicit time–frequency representation learning.
The branch network adopts a stage-wise hierarchical architecture, in which WTConv blocks and standard convolution blocks are alternately stacked. Each convolution block consists of a conventional convolution layer followed by batch normalization (BN) \cite{ioffe2015batch} and a Leaky ReLU activation, and is primarily responsible for spatial–temporal feature extraction and progressive downsampling. Each WTConv block is composed of a wavelet convolution layer followed by BN and a Leaky ReLU activation, aiming to explicitly encode multi-scale frequency information at the corresponding resolution level.
Fig.~\ref{img:WTconv} outlines the process of the WTConv layer.

In the WTConv layer, the input feature map is first decomposed via a two-dimensional discrete wavelet transform (WT) to achieve multi-resolution representation. We adopt the Haar wavelet basis, whose direction-sensitive filter bank decomposes the input into one low-frequency approximation component and three high-frequency detail components. This convolutional wavelet transform enables the network to explicitly separate different frequency bands in seismic data and capture frequency-dependent structural characteristics.

After multi-scale decomposition, convolution operations are independently applied to features in different frequency bands, facilitating the extraction of global low-frequency information and directional high-frequency details. The processed multi-band features are then fused through an inverse wavelet transform (IWT) and combined with the input through a residual connection, yielding the final output of the WTConv layer.

\subsubsection{Source-parameterized trunk network}
In seismic inversion, the source is not only the origin of wavefield propagation but also a key physical factor affecting imaging quality and inversion stability. The source excitation, spatial location, and frequency characteristics determine the energy distribution of seismic waves in both time and frequency domains, thereby directly influencing the sensitivity of observations to subsurface structures at different scales.

To explicitly incorporate these controllable physical factors, we design a source-parameterized trunk network that models seismic sources in the parameter space rather than the conventional spatial coordinate domain \cite{zhu2023fourier}. Specifically, the physical source parameters, including frequency and location, are taken as the inputs to the trunk network. This design enables the network to learn high-dimensional latent representations driven by source physics, facilitating improved generalization across varying source conditions.
The trunk network is implemented as a MLP composed of several fully connected layers, each followed by a ReLU activation function.

% We take source physical features (frequency and location) as the input of trunk net.
% In other words, we establish Inversion-DeepONet in the source parameter space instead of traditional spatial domain \citep{zhu2023fourier}.
% %There also have another advantage: decrease the amount of training data.
% Additionally, this approach offers the advantage of reducing the amount of training data.
% According to the definition of vanilla DeepONet and Equation~\eqref{acoustic wave equation}, the coordinates ($x$, $z$) of final output $c$ are fed to trunk net.
% However, there are lots of coordinates in one velocity model $c$, and every coordinate will be processed through trunk net for computation.
% This issue will be optimized when taking source parameters as the input of trunk net.
Then, the inversion mapping of FWI with varying parameters can be demonstrated as follows:
\begin{equation}\label{eq:inversion2}
c=g^{-1}(p, \boldsymbol{\xi}),
\end{equation}
where source parameter $\boldsymbol{\xi}$ is the variable-length vector, and it has a length of 5 indicating the source locations or frequencies of five sources, or has a length of 10 indicating both source locations or frequencies.

\subsubsection{Interactive branch–trunk decoding network}

The branch network acts as an encoder that progressively abstracts seismic observations using WTConv and CNNs, extracting multi-scale time–frequency features from raw waveforms. In parallel, the trunk network performs a nonlinear mapping of source parameters, producing a high-dimensional latent representation conditioned on source physics.

The latent features from the branch and trunk networks are interactively fused through element-wise multiplication:
\begin{equation}
h=b \odot t,
\end{equation}
where $b$ and $t$ denote the output feature vectors of the branch and trunk networks, respectively, and $\odot$ represents the Hadamard product. This fused representation jointly encodes seismic observations and source-dependent physical information, serving as the foundational feature for velocity model reconstruction.

To map the fused latent features back to the spatial domain, a decoder based on transposed convolutional neural networks is employed at the network output, progressively recovering the two-dimensional velocity model.

\section{Experiments}

In this section, we demonstrate the accuracy and generalization ability of SG-DeepONet on our proposed SVFWI dataset. 
Specifically, we test the proposed method on three types of datasets (SVFWI-F, SVFWI-L, SVFWI-FL), compared with three baseline models: vanilla DeepONet ,InversionNet ,and Fourier-DeepONet.
Then, We evaluate the generalization ability across different source-condition scenarios.

% \begin{table}[h]
% \caption{The Data partitioning of FWI-F of FWI-F, FWI-L and FWI-FL. }
% \label{table:data}
% % \setlength\tabcolsep{0.4cm}
% \centering
% \begin{tabular}{llrr} 
% \hline
% \multirow{2}*{Datasets} & \multirow{2}*{Subsurface structure} & Number of  & Number of   \\
% ~ & ~ & training set & testing set \\\hline
% \multirow{4}*{FWI-F} & FlatVel-B & 24,000 & 6,000  \\
% ~ & CurveVel-A & 24,000 & 6,000 \\
% ~ & FlatFault-B & 48,000 & 6,000 \\
% ~ & CurveFault-A & 48,000 & 6,000 \\\hline
% \multirow{4}*{FWI-L} & FlatVel-B & 24,000 & 6,000  \\
% ~ & CurveVel-A & 24,000 & 6,000 \\
% ~ & FlatFault-B & 48,000 & 6,000 \\
% ~ & CurveFault-A & 48,000 & 6,000 \\\hline
% FWI-FL &  CurveVel-A & 24,000 & 6,000  \\\hline
% \end{tabular}
% \end{table}

\subsection{Implementation details}
\paragraph{Data description}
The neural network aims at learning the mapping from seismic data $p \in \mathbb{R}^{S\times T \times R}$ to velocity model $c \in \mathbb{R}^{W\times H} $, where $W$ and $H$ denote the horizontal and vertical dimensions of the velocity model, $S$ represents the number of seismic sources, $T$ indicates the frequency of receiver recording in one second, and $R$ represents the total number of receivers.
In this work, the dimensions of seismic data and velocity model are all $5\times 1000 \times 70$ and $70 \times 70 $.
Each dataset represents a 0.7km × 0.7km subsurface area discretized into 70×70 pixels, with seismic wave velocities varying approximately from 1.5 km/s to 4.5 km/s.
We employ five seismic sources and capture 1,000 sensor readings of the resulting wavefield within a one‑second window.
The partitioning of the datasets into training and testing sets follows the protocol established by OpenFWI \cite{deng2022openfwi}.

\paragraph{Training setting}
In our experiments, we set the batch size to 128 and total number of training epochs is set to 120.
The learning rate is initialized to 0.001 and decayed by a factor of 0.9 every 20 epochs.
The convolutional layers are initialized with the Kaiming uniform scheme, whereas the fully connected layers employ Xavier initialization to maintain well‑conditioned signal propagation throughout the network.
We utilize AdamW optimizer with momentum parameters $\beta_1 = 0.9$, $\beta_2 = 0.999$ and a weight decay of $1 \times 10^{-4}$ to update all parameters of the network.
In addition, all the seismic data and velocity models are normalized between -1 and 1.
All experiments are performed on NVIDIA Tesla V100 32 GB, using the PyTorch GPU framework.

\paragraph{Evaluation metrics}
%To evaluate the accuracy of network models, 
We utilize several metrics: mean absolute error (MAE), root mean square error (RMSE) \cite{deng2022openfwi}, structural similarity (SSIM) \cite{wang2004image}, and relative error (RE).
The MAE, RMSE, and RE are introduced to determine the accuracy of numerical values in velocity models.
SSIM is used to evaluate the quality of images by measuring the similarity in structure and content between two images.
The MAE, RMSE and RE focus on the pixel difference, and SSIM focus on the whole.
The MAE, RMSE and RE can be represented as follows:
\begin{equation}\label{eq:RE}
\text{MAE} = \frac{1}{N} \sum_{i=1}^N \left (\frac{1}{M}\sum_{j=1}^M {|c_{j}^i-\hat{c}_{j}^i|} \right),
\end{equation}

\begin{equation}\label{eq:RE}
\text{RMSE} = \frac{1}{N} \sum_{i=1}^N \sqrt{ \frac{1}{M} \sum_{j=1}^M {|c_{j}^i-\hat{c}_{j}^i|}^2},
\end{equation}

\begin{equation}\label{eq:RE}
\text{RE} = \frac{1}{N} \sum_{i=1}^N \sqrt{\frac{\sum_{j=1}^M {|c_{j}^i-\hat{c}_{j}^i|}^2}{\sum_{j=1}^M {|c_{j}^i|}^2}},
\end{equation}
where $N$ is the number of all velocity models used to test, $M$ is the number of pixels in one velocity model. 
The prediction is denoted as $\hat{c}$, and the ground truth is denoted as $c$.
We take MAE as the loss function for all experiments.

\begin{figure*}
   \centering
    \includegraphics[width=0.95\textwidth]{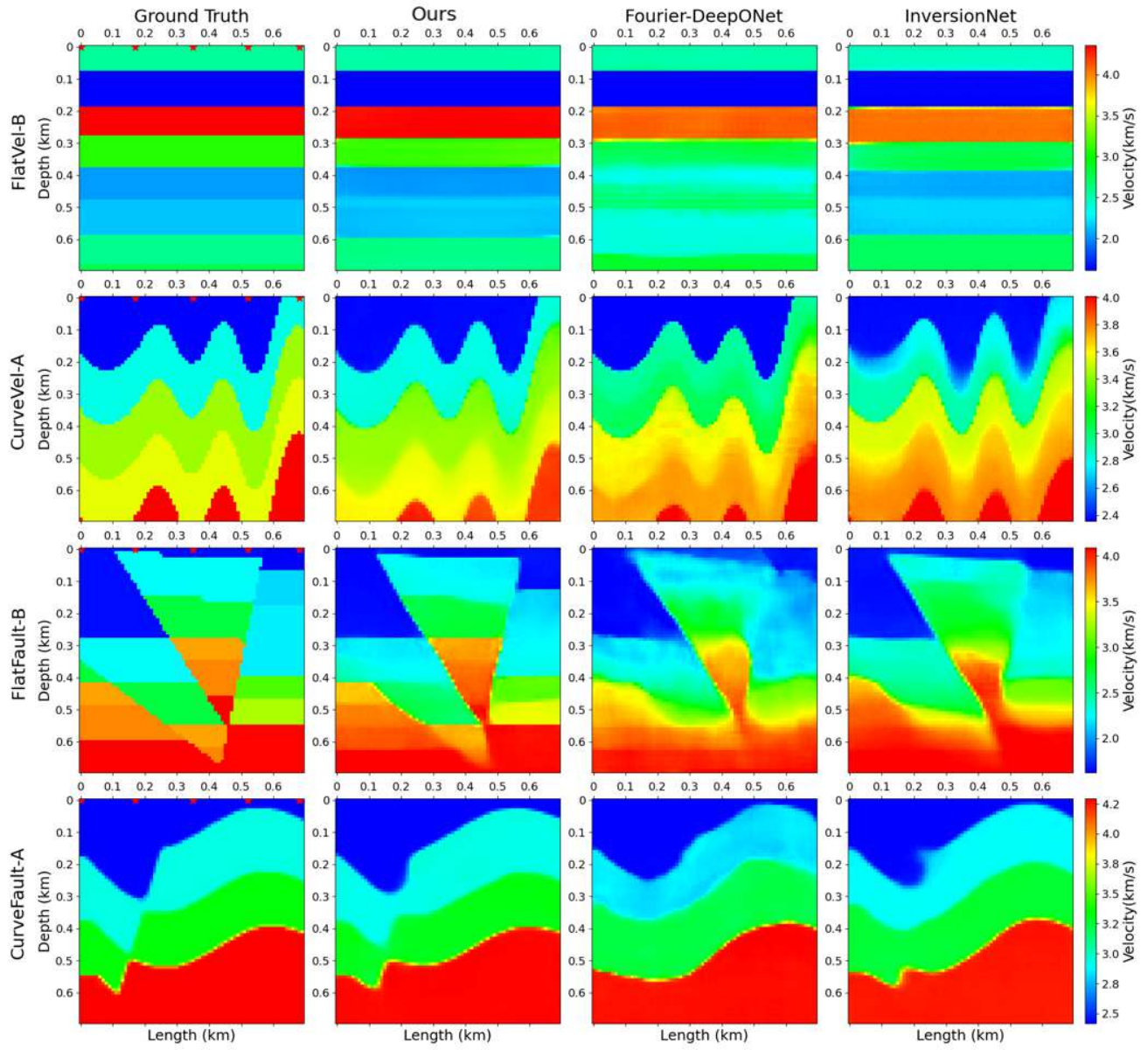}
     \caption{Examples of inversion results predicted by InversionNet-DeepONet, Fourier-DeepONet and
InversionNet tested on four categories of velocity models from SVFWI-F. The five red stars in the ground truth
are seismic sources, which have different frequencies and fixed location.}
     \label{fig:FWI-F}
\end{figure*}

\begin{table*}[h]
\caption{The quantitative results of vanilla DeepONet , InversionNet, Fourier-DeepONet and SG-DeepONet tested on SVFWI-F. 
% We validate these methods on four categories of velocity models, where five sources have fixed locations and different frequencies.}
%between 5 and 25 Hz. The Inversion-DeepONet has the best performance under all metrics. 
The five sources in the four dataset categories have fixed spatial locations but vary in frequency within the range of 5 to 25 Hz. 
}
\label{table:FWI-F}
\setlength\tabcolsep{0.45cm} 
\centering
\begin{tabular}{llrrrr} \hline
Dataset & Model & MAE $\downarrow$ & RMSE $\downarrow$ & SSIM $\uparrow$ & RE $\downarrow$ \\\hline
\multirow{4}*{FlatVel-B} & Vanilla DeepONet & 0.1031 & 0.2005 & 0.7834 & 0.2995 \\
~ & InversionNet & 0.0781 & 0.1833 & 0.8354 & 0.2756  \\
~ & Fourier-DeepONet & 0.0637 & 0.1564 & 0.8602 & 0.2281 \\
~ & \textbf{Ours} & \textbf{0.0501} & \textbf{0.1395} & \textbf{0.8868} & \textbf{0.1955} \\\hline
\multirow{4}*{CurveVel-A} & Vanilla DeepONet & 0.1100 & 0.1652 & 0.7248 & 0.3141\\
~ & InversionNet & 0.0861 & 0.1479 & 0.7596 & 0.2697 \\
~ & Fourier-DeepONet & 0.0717 & 0.1326 & 0.7984 & 0.2393 \\
~ & \textbf{Ours} & \textbf{0.0580} & \textbf{0.1179} & \textbf{0.8273} & \textbf{0.2118} \\\hline
\multirow{4}*{FlatFault-B} & Vanilla DeepONet & 0.1377 & 0.2198 & 0.6521 & 0.3533 \\ 
~ & InversionNet & 0.1081 & 0.1706 & 0.7048 & 0.2935 \\
~ & Fourier-DeepONet & 0.1057 & 0.1648 & 0.7122 & 0.2871 \\
~ & \textbf{Ours} & \textbf{0.0983} & \textbf{0.1619} & \textbf{0.7161} & \textbf{0.2790} \\\hline
\multirow{4}*{CurveFault-A} & Vanilla DeepONet & 0.0796 & 0.1389 & 0.8379 & 0.1771 \\
~ & InversionNet & 0.0364 & 0.0883 & 0.9285 & 0.1493 \\
~ & Fourier-DeepONet & 0.0411 & 0.0978 & 0.9152 & 0.1670 \\
~ & \textbf{Ours} & \textbf{0.0266} & \textbf{0.0758} & \textbf{0.9414} & \textbf{0.1249} \\\hline
\end{tabular}

\end{table*}

The SSIM measures perceived image quality by comparing luminance, contrast, and structural information between a reference image $x$ and a test image $y$. 
Unlike pixel‑wise metrics, SSIM models properties of the human visual system, yielding scores that align more closely with subjective judgments.
For a local window, SSIM is defined as:
\begin{equation}
 S S I M ( x , y ) = \left[ l ( x , y ) \right] ^ { \alpha } \left[ c ( x , y ) \right] ^ { \beta } \left[ s ( x , y ) \right] ^ { \gamma },
\end{equation}

% \begin{equation}
% \left\{
% \begin{array}{l}
%  l ( x , y ) = \frac { 2 \mu _ { x } \mu _ { y } + C _ { 1 } } { \mu _ { x } ^ { 2 } + \mu _ { y } ^ { 2 } + C _ { 1 } } ,\\
%  c ( x , y ) = \frac { 2 \sigma _ { x } \sigma _ { y } + C _ { 2 } } { \sigma _ { x } ^ { 2 } + \sigma _ { y } ^ { 2 } + C _ { 2 }} ,\\
% s ( x , y ) = \frac { \sigma _ { x y } + C _ { 3 } } { \sigma _ { x } \sigma _ { y } + C _ { 3 } }.
% \end{array}
% \right.
% \end{equation}

\begin{equation}
\begin{cases}
    l ( x , y ) = \frac { 2 \mu _ { x } \mu _ { y } + C _ { 1 } } { \mu _ { x } ^ { 2 } + \mu _ { y } ^ { 2 } + C _ { 1 } } ,\\
     c ( x , y ) = \frac { 2 \sigma _ { x } \sigma _ { y } + C _ { 2 } } { \sigma _ { x } ^ { 2 } + \sigma _ { y } ^ { 2 } + C _ { 2 }} ,\\
s ( x , y ) = \frac { \sigma _ { x y } + C _ { 3 } } { \sigma _ { x } \sigma _ { y } + C _ { 3 } },
\end{cases}
\end{equation}
where the  $\mu _ { x } , \mu _ { y }$ are the local means of $x$ and $y$.  
$\sigma _ { x } , \sigma _ { y }$ are the local standard deviations and  $\sigma _ { x y }$ is the local covariance.
The constants $C_{1}, C_{2}, C_{3}$ act as stability terms to avoid division by zero, and an SSIM value approaching 1 signifies that the two images are more similar.

\subsection{Results on varying source frequencies}\label{sec:FWI-F}

% \begin{figure*}[h]
%    \centering
%     \includegraphics[width=0.9\textwidth]{FIG/CV-FL.png}
%      \caption{The predictions of three models tested on CurveVel-A in FWI-FL. For this dataset, The frequency and horizontal location of the five sources are all varying within a certain range.}
%      \label{fig:FWI-FL}
% \end{figure*}
% \begin{table*}[h]
% \setlength\tabcolsep{0.4cm}
% \centering
% \begin{tabular}{ccccccc} 
% \hline
% Encoder-Decoder & MAE $\downarrow$ & RMSE $\downarrow$ & SSIM $\uparrow$ & RE $\downarrow$& Batch Size & Speed (s/epoch) \\\hline
% InverisonNet & \textbf{0.0591} & \textbf{0.1195} & \textbf{0.8262} & \textbf{0.2136} & 128 & 62 \\
% Fourier-DeepONet & 0.0717 & 0.1326 & 0.7984 & 0.2393 & 32 & 525 \\
% DD-Net70 & 0.0828 & 0.1409 & 0.7749 & 0.2558 & 128 & 50 \\
% Vanilla DeepONet & 0.1100 & 0.1652 & 0.7248 & 0.3141 & 128 & 53 \\\hline
% \end{tabular}
% \caption{The quantitative results of the four encoder-decoder structures tested on CurveVel-A in FWI-F. The encoder-decoder of InversionNet, Fourier-DeepONet and DD-Net70 are employed in novel DeepONet architecture. For DD-Net70, we just use the first decoder. For Vanilla DeepONet, the coordinate matrix of velocity model are fed into trunk net.}
% \label{table:curvevel-A-F}
% \end{table*}

%The sources with different frequencies play an important role in FWI. 
Low frequencies provide deeper penetration and are less affected by noise, which helps in resolving larger-scale structures and provides information about the deeper parts of the subsurface.
High Frequencies offer higher resolution and are more sensitive to smaller-scale features which help in imaging finer details of the subsurface.
Therefore, using a range of frequencies can avoid local minima improving  the convergence and allow for a more detailed and accurate reconstruction of the subsurface properties.

We take vanilla DeepONet, InversionNet~\cite{wu2019inversionnet} ,and Fourier-DeepONet~\cite{zhu2023fourier} as three baseline models.
We validate our method on SVFWI-F where five sources have fixed locations and different frequencies varying from 5 to 25 Hz.
Table~\ref{table:FWI-F} demonstrates the quantitative results on four categories of velocity models.
Our method consistently shows the lowest error and highest SSIM.
Fig.~\ref{fig:FWI-F} presents the inversion results of three models.
In FlatVel-B and CurveVel-A, our method can outline the clear boundaries and provide accurate velocity.
However, Fourier-DeepONet and InversionNet predict wrong velocity in some layers and generate vague interfaces.
In FlatFault-B and CurveFault-A, although the inversion results of SG-DeepONet have some deviation from the ground truth, compared with the baseline models, the fault position and shape have been basically identified. 

%Specifically, the predictions of Fourier-DeepONet and InversionNet in CurveFault-A without fault in green layer (row 4 in Figure \ref{fig:FWI-F}) are obvious different from the ground truth.

% \begin{table*}[htb]
% \caption{The quantitative results of the four encoder-decoder structures tested on CurveVel-A from FWI-F. The encoder-decoder of InversionNet, Fourier-DeepONet, and DD-Net70 are employed in our novel DeepONet architecture. For DD-Net70, we just use the first decoder. For Vanilla DeepONet, the coordinate matrix of velocity model are fed into trunk net.}
% \label{table:curvevel-A-F}
% \setlength\tabcolsep{0.4cm}
% \centering
% \begin{tabular}{lrrrrrr} 
% \hline
% Encoder-Decoder & MAE $\downarrow$ & RMSE $\downarrow$ & SSIM $\uparrow$ & RE $\downarrow$& Batch Size & Speed (s/epoch) \\\hline
% InverisonNet & \textbf{0.0591} & \textbf{0.1195} & \textbf{0.8262} & \textbf{0.2136} & 128 & 62 \\
% Fourier-DeepONet & 0.0717 & 0.1326 & 0.7984 & 0.2393 & 32 & 525 \\
% DD-Net70 & 0.0828 & 0.1409 & 0.7749 & 0.2558 & 128 & 50 \\
% Vanilla DeepONet & 0.1100 & 0.1652 & 0.7248 & 0.3141 & 128 & 53 \\\hline
% \end{tabular}

% \end{table*}

\begin{figure*}
   \centering
    \includegraphics[width=0.9\textwidth]{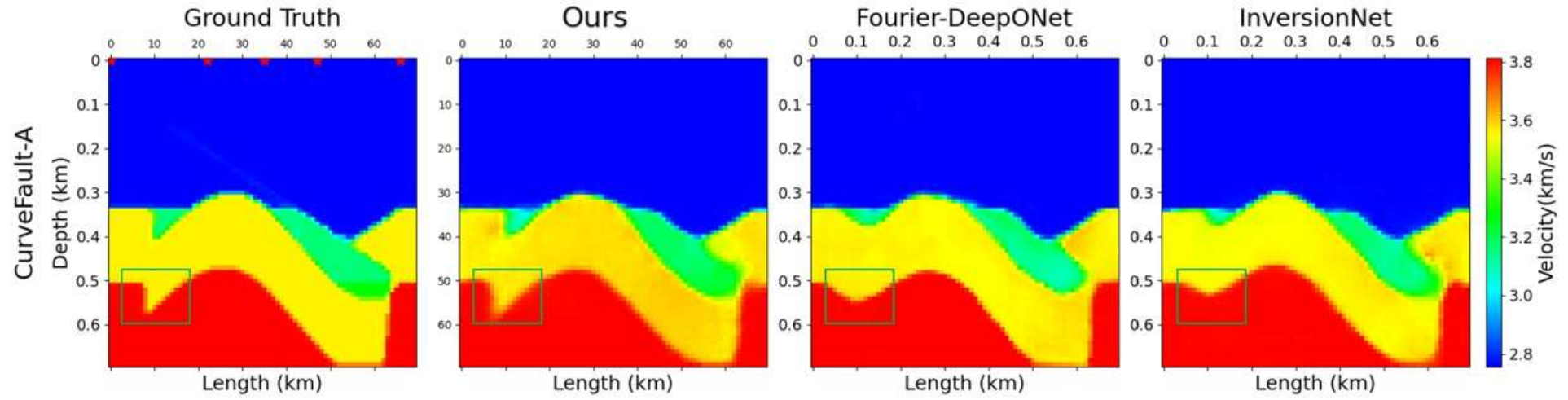}
     \caption{The predictions of three models tested on CurveFault-A from SVFWI-L. For this example, the five source locations are 1.8, 222.7, 359.9, 477.2, and 662.8 m (red stars in ground truth). Additionally, all five source frequencies are uniformly set at 15 Hz.}
     \label{fig:FWI-L}
\end{figure*}

\begin{table*}[tbp]
\caption{The quantitative results of vanilla DeepONet, InversionNet, Fourier-DeepONet and SG-DeepONet tested on SVFWI-L. The five sources in four categories of datasets have fixed frequency at 15 Hz and varying locations in [0, 50], [122.5, 222.5], [295, 395], [467.5, 567.5] and [640, 690] m. SG-DeepONet achieves the best performance on three of the datasets.}
\label{table:FWI-L}
\setlength\tabcolsep{0.5cm} 
\centering
\begin{tabular}{llrrrr} \hline
Dataset & Model & MAE $\downarrow$ & RMSE $\downarrow$ & SSIM $\uparrow$ & RE $\downarrow$ \\\hline
\multirow{4}*{FlatVel-B} & Vanilla DeepONet & 0.0581 & 0.1174 & 0.8913 & 0.1534\\
~ & InversionNet & 0.0354 & 0.0911 & 0.9445 & 0.1179  \\
~ & Fourier-DeepONet & 0.0286 & 0.0911 & 0.9442 & 0.1067 \\
~ & \textbf{Ours} & \textbf{0.0256} & \textbf{0.0744} & \textbf{0.9588} & \textbf{0.0872} \\\hline
\multirow{4}*{CurveVel-A} & Vanilla DeepONet & 0.0816 & 0.1394 & 0.7816 & 0.2537 \\
~ & InversionNet & 0.0541 & 0.1123 & 0.8351 & 0.2025  \\
~ & Fourier-DeepONet & 0.0499 & 0.1051 & 0.8522 & 0.1892  \\
~ & \textbf{Ours} & \textbf{0.0374} & \textbf{0.0941} & \textbf{0.8731} & \textbf{0.1701}  \\\hline
\multirow{4}*{FlatFault-B} & Vanilla DeepONet & 0.1159 & 0.1903 & 0.6974 & 0.3127 \\
~ & InversionNet & 0.0859 & 0.1525 & 0.7488 & 0.2615  \\
~ & Fourier-DeepONet & 0.0787 & 0.1448 & 0.7826 & 0.2462  \\
~ & \textbf{Ours} & \textbf{0.0754} & \textbf{0.1403} & \textbf{0.7838} & \textbf{0.2402}  \\\hline
\multirow{4}*{CurveFault-A} & Vanilla DeepONet & 0.0476 & 0.1045 & 0.8987 & 0.1773 \\ 
~ & InversionNet & 0.0223 & 0.0622 & 0.9564 & 0.1039  \\
~ & Fourier-DeepONet & 0.0219 & \textbf{0.0613} & \textbf{0.9575} & \textbf{0.0989}  \\
~ & \textbf{Ours} & \textbf{0.0211} & 0.0640 & 0.9521 & 0.1071  \\\hline
\end{tabular}

\end{table*}

\subsection{Results on varying source locations}\label{sec:FWI-L}
%The source locations also play a crucial role in FWI.
Variable-location sources can make more areas effectively investigated, providing more comprehensive information.
In addition, it can collect data from different angles, generating a reliable imaging of the subsurface.
We test four models on SVFWI-L, which have fixed source frequency at 15 Hz and varying source locations within a certain range.
We select FlatVel-B, CurveVel-A, FlatFault-B, and CurveFault-A from SVFWI-L for testing purposes.
%Figure \ref{fig:FWI-L} shows the inversion results of three methods on FWI-L, and the prediction of IversionNet-DeepONet closely resemble the ground truth.
%The quantitative results of three models are shown in Table \ref{table:FWI-L} and Inversion-DeepONet exhibits the best performance on three of the datasets.
Table~\ref{table:FWI-L} demonstrates the quantitative
results on four categories of velocity models. 
Our method consistently shows the lowest error and the highest SSIM on FlatVel-B, CurveVel-A and FlatFault-B.
The predicted velocity models of three models are shown in Fig.~\ref{fig:FWI-L}. As highlighted by the green box, our method achieves substantially more accurate inversion result.

However, compared with experiment on varying source frequencies (SVFWI-F), the numerical differences of four models are not such obvious.
On the one hand, learning the inversion mapping in SVFWI-L is relatively easier than SVFWI-F.
On the other hand, the source locations are implicitly embedded in the seismic data.
As can be seen, the peak of the waveform in seismic data (the input of branch net in Fig.~\ref{img:architecture}) corresponds to the source location, and five seismic data channels stem from five sources.
For this reason, InversionNet can gain the information about source locations without the input of  parameters vector, which is fed to trunk net in SG-DeepONet and Fourier-DeepONet. 
Our method achieves the best performance on most datasets in SVFWI-L.
%Hence, those three models show a comparable performance in FWI-L.  

\subsection{Results on varying source frequencies and locations}\label{sec:FWI-FL}

\begin{figure*}
   \centering
    \includegraphics[width=0.9\textwidth]{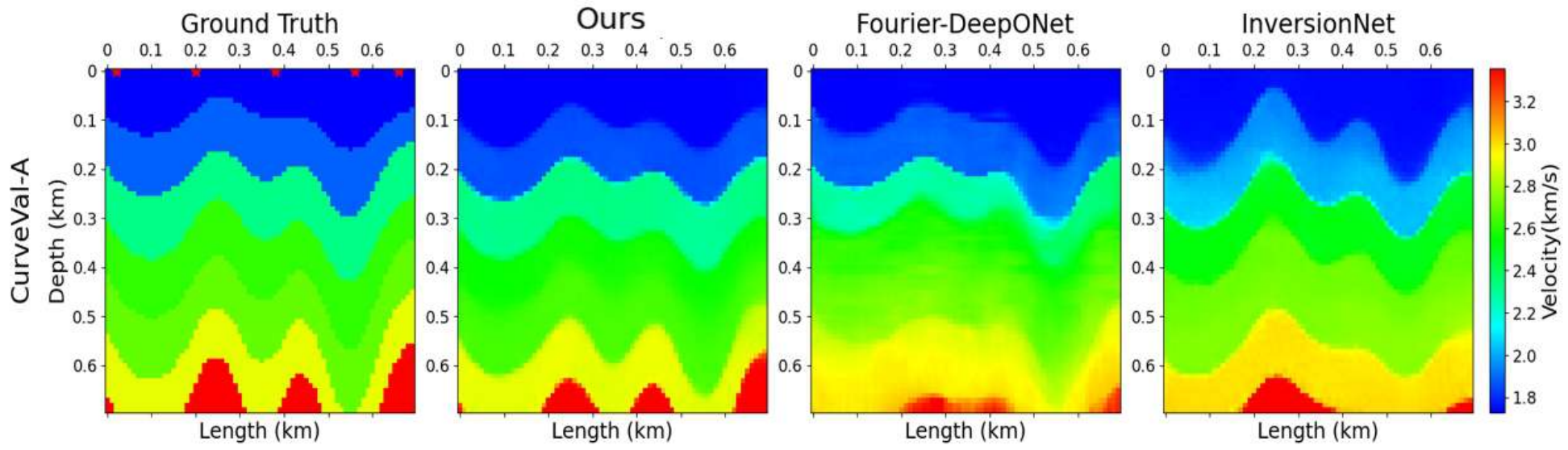}
     \caption{The predictions of three models tested on CurveVel-A in SVFWI-FL. For this example, the five source locations are 27.8, 201.7, 384.9, 565.2 and 666.8 m (red stars in ground truth). And the five source frequencies are 6.3, 15.8, 8.6, 22.0 and 7.9 Hz.}
     \label{fig:FWI-FL}
\end{figure*}

We introduce the advantages of using varying source frequencies or locations in FWI above, respectively.
This section, we test the performance of four models on CurveVel-A from SVFWI-FL, which has both varying source frequencies and locations.
SVFWI-FL combines the advantages of both SVFWI-F and SVFWI-L. 
However, it also increases the learning complexity.
The quantitative results of four models are presented in Table~\ref{table:FWI-FL}.
%and the visualization images are displayed in Figure~\ref{fig:FWI-FL}. 
%and the inversion results of three models are shown in Figure \ref{fig:FWI-FL}.
Obviously, our method outperforms those baseline models and achieves results that better matches the ground truth.
The visual results in Fig.~\ref{fig:FWI-FL} are consistent with the numerical results.

\begin{table}
\caption{The quantitative results of the four models tested on CurveVel-A from SVFWI-FL.}
\label{table:FWI-FL}
\centering
\begin{tabular}{lrrrr} 
\hline
Model & MAE$\downarrow$ & RMSE$\downarrow$ & SSIM$\uparrow$ & RE$\downarrow$  \\\hline
Vanilla DeepONet & 0.1263 & 0.1927 & 0.7033 & 0.3185 \\
InverisonNet & 0.0888 & 0.1511 & 0.7544 & 0.2762 \\
Fourier-DeepONet & 0.0773 & 0.1383 & 0.7884 & 0.2523 \\
\textbf{Ours} & \textbf{0.0677} & \textbf{0.1293} & \textbf{0.8012} & \textbf{0.2363}  \\\hline
\end{tabular}

\end{table}

\subsection{Generalization analysis}\label{sec:generalization}
A model with good generalization ability can maintain stable performance when faced with different types of seismic data or varying conditions, demonstrating stronger robustness.
In practical applications, acquiring seismic data and velocity model is costly and difficult. 
Hence, good generalization ability has a great impact on FWI, making it more practical and valuable.

\begin{figure*}
   \centering
    \includegraphics[width=0.95\textwidth]{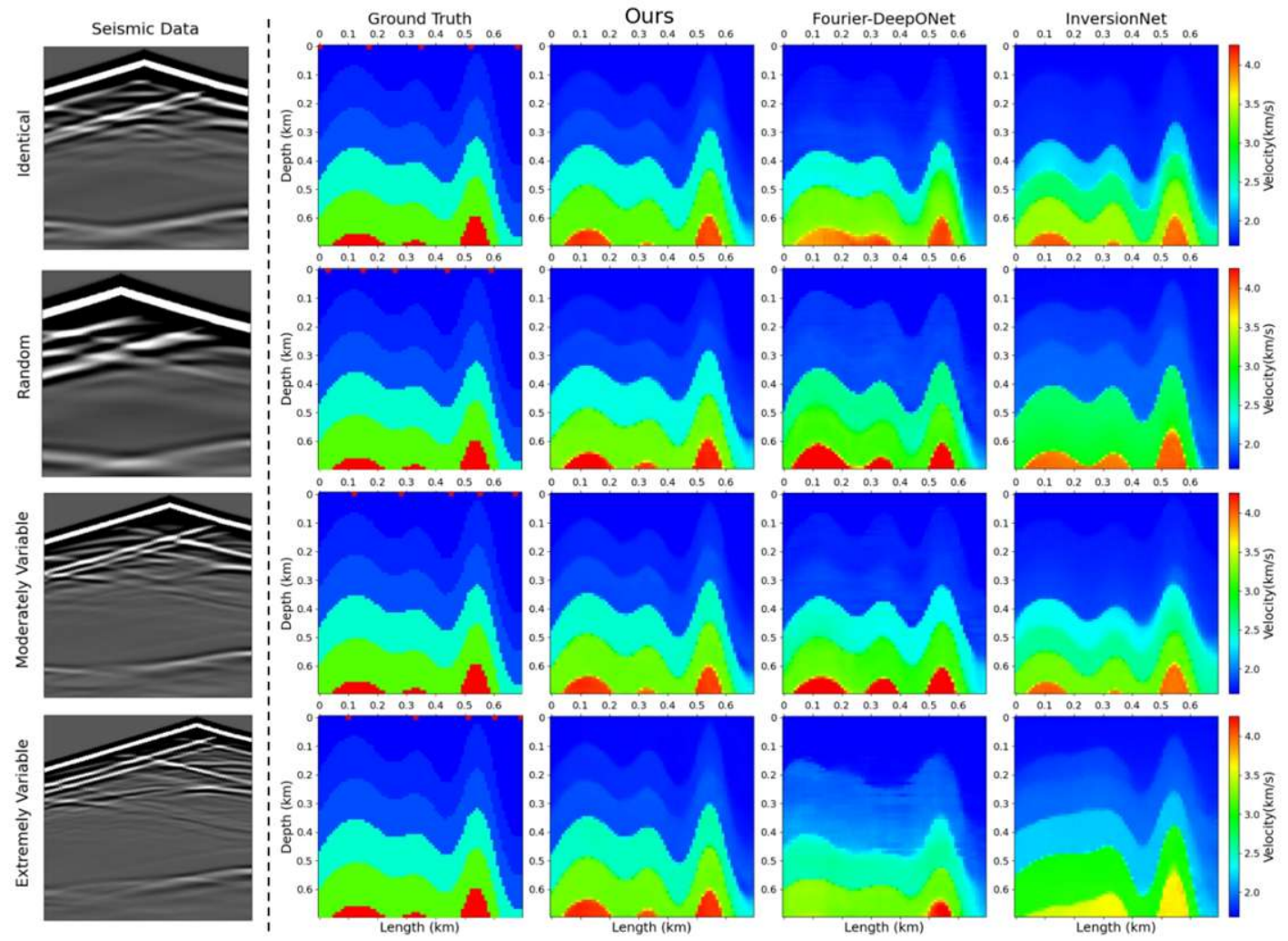}
     \caption{The generalization ability of source parameters between SG-DeepONet, Fourier-DeepONet and InversionNet tested on CurveVel-A in SVFWI-FL. 
     On the left side are seismic data from Source C, which serve as part of the input of neural network.
     The seismic data are generated by sources with varying frequencies and locations.
     Four scenarios of five source frequencies and locations are: 
      (1) Identical. (2) Random within a certain range.  (3) Moderately variable within a small range. (4) Extremely variable within a large range.}
     %Examples of different scenarios of five source frequencies: (1) Identical at 15 Hz. (2) Random at 19.9, 9.7, 10.0, 16.4 and 5.2 Hz. (3) Moderately variable at 10, 15, 20, 15 and 10 Hz. (4) Extremely variable at 5, 15, 25, 15 and 5 Hz. }
     \label{fig:generalization}
\end{figure*}

In this section, we compare the generalizability of seismic sources between our method and the baseline models.
We utilize a single velocity model to generate various types of seismic data with different source frequencies and locations based on Eq.~\eqref{acoustic wave equation}.
Subsequently, we evaluate the performance of three parameterized models, pre-trained on CurveVel-A from SVFWI-FL, on seismic data under different scenarios. 
These scenarios include cases where the source frequencies and locations of five sources are identical, random, moderately variable, or extremely variable.
As shown in Fig.~\ref{fig:generalization}, regardless of the variations in source parameters, the inversion results of SG-DeepONet remain nearly unchanged and are the most consistent with the ground truth.
In contrast, the predictions of InversionNet exhibit significant changes when source frequencies and locations vary, particularly in extreme cases (5 Hz or 25 Hz). 
Moreover, while Fourier-DeepONet is less sensitive to changes in source parameters compared to InversionNet, its accuracy is still inferior to that of our proposed method.

\section{Conclusion}
In this study, we construct a new seismic dataset, termed SVFWI, to address the source generalization limitations of OpenFWI.
SVFWI explicitly models source diversity by varying source frequencies and locations across multiple surface sources, providing a more challenging benchmark for data-driven full waveform inversion.
Moreover, we propose SG-DeepONet, a DeepONet-based encoder–decoder framework.
% In this study, we address the limitations of OpenFWI and generate a more promising and potential dataset. 
% The enhanced dataset, which incorporates five sources spanning both low and high frequencies, captures comprehensive subsurface information.  
% In addition, we propose a novel DeepONet architecture with encoder-decoder structure, termed SG-DeepONet, for solving seismic imaging inverse problem.
Our method incorporates prior knowledge of the seismic source into the neural network, and adopts an encoder-decoder architecture to more effectively reconstruct the velocity model.
Experimental results show that our method exhibits a obviously improved accuracy and generalization ability.

In our current work, the proposed method is a purely data-driven supervised learning approach, which relies on large amounts of labeled data. However, such data is often difficult to obtain in real-world scenarios. For future work, we aim to incorporate prior physical constraints—such as the acoustic wave partial differential equation—into the model training process. This would enable the development of an unsupervised or semi-supervised learning framework that is more applicable to real-world settings.

%Bibliography
\bibliographystyle{unsrt}  
\bibliography{references}

@article{wang2004image,
  title={Image quality assessment: from error visibility to structural similarity},
  author={Wang, Z. and Bovik, A. C. and Sheikh, H. R. and Simoncelli, E. P.},
  journal={IEEE transactions on image processing},
  volume={13},
  number={4},
  pages={600--612},
  year={2004},
  publisher={IEEE}
}

@article{virieux2009overview,
  title={An overview of full-waveform inversion in exploration geophysics},
  author={Virieux, J. and Operto, S.},
  journal={Geophysics},
  volume={74},
  number={6},
  pages={WCC1--WCC26},
  year={2009},
  publisher={Society of Exploration Geophysicists}
}

@article{guasch2020full,
  title={Full-waveform inversion imaging of the human brain},
  author={Guasch, L. and Calder{\'o}n, A. O. and Tang, M. X. and Nachev, P. and Warner, M.},
  journal={NPJ digital medicine},
  volume={3},
  number={1},
  pages={28},
  year={2020},
  publisher={Nature Publishing Group UK London}
}

@inproceedings{zhang2020fwi,
  title={FWI Imaging: Full-wavefield imaging through full-waveform inversion},
  author={Zhang, Z. G. and Wu, Z. D. and Wei, Z. Y. and Mei, J. W. and Huang, R. X. and Wang, P.},
  booktitle={SEG International Exposition and Annual Meeting},
  pages={D031S027R004},
  year={2020},
  organization={SEG}
}

@article{lu2021learning,
  title={Learning nonlinear operators via DeepONet based on the universal approximation theorem of operators},
  author={Lu, L. and Jin, P. Z. and Pang, G. F. and Zhang, Z. Q. and Karniadakis, G. E.},
  journal={Nature machine intelligence},
  volume={3},
  number={3},
  pages={218--229},
  year={2021},
  publisher={Nature Publishing Group UK London}
}

@inproceedings{ronneberger2015u,
  title={U-net: Convolutional networks for biomedical image segmentation},
  author={Ronneberger, O. and Fischer, P. and Brox, T.},
  booktitle={Medical image computing and computer-assisted intervention--MICCAI 2015: 18th international conference, Munich, Germany, October 5-9, 2015, proceedings, part III 18},
  pages={234--241},
  year={2015},
  organization={Springer}
}

@misc{li2020fourier,
  title={Fourier neural operator for parametric partial differential equations},
  author={Li, Z. Y. and Kovachki, N. and Azizzadenesheli, K. and Liu, B. and Bhattacharya, K. and Stuart, A. and Anandkumar, A.},
  year={2020},
  eprint={2010.08895},
  archivePrefix={arXiv}
}

@article{deng2022openfwi,
  title={OpenFWI: Large-scale multi-structural benchmark datasets for full waveform inversion},
  author={Deng, C. Y. and Feng, S. H. and Wang, H. C. and Zhang, X. T. and Jin, P. and Feng, Y. N. and Zeng, Q. L. and Chen, Y. P. and Lin, Y. Z.},
  journal={Advances in Neural Information Processing Systems},
  volume={35},
  pages={6007--6020},
  year={2022}
}

@article{zhu2023fourier,
  title={Fourier-DeepONet: Fourier-enhanced deep operator networks for full waveform inversion with improved accuracy, generalizability, and robustness},
  author={Zhu, M. and Feng, S. H. and Lin, Y. Z. and Lu, L.},
  journal={Computer Methods in Applied Mechanics and Engineering},
  volume={416},
  pages={116300},
  year={2023},
  publisher={Elsevier}
}

@article{mei2024fully,
  title={Fully Convolutional Network enhanced DeepONet-based surrogate of predicting the travel-time fields},
  author={Mei, Y. F. and Zhang, Y. J. and Zhu, X. Y. and Gou, R. X. and Gao, J. H.},
  journal={IEEE Transactions on Geoscience and Remote Sensing},
  year={2024},
  publisher={IEEE}
}

@inproceedings{wang2018velocity,
  title={Velocity model building with a modified fully convolutional network},
  author={Wang, W. L. and Yang, F. S. and Ma, J. W.},
  booktitle={SEG International Exposition and Annual Meeting},
  pages={SEG--2018},
  year={2018},
  organization={SEG}
}

@article{wu2019inversionnet,
  title={InversionNet: An efficient and accurate data-driven full waveform inversion},
  author={Wu, Y. and Lin, Y. Z.},
  journal={IEEE Transactions on Computational Imaging},
  volume={6},
  pages={419--433},
  year={2019},
  publisher={IEEE}
}

@inproceedings{zhang2019velocitygan,
  title={VelocityGAN: Subsurface velocity image estimation using conditional adversarial networks},
  author={Zhang, Z. P. and Wu, Y. and Zhou, Z. and Lin, Y. Z.},
  booktitle={2019 IEEE Winter Conference on Applications of Computer Vision (WACV)},
  pages={705--714},
  year={2019},
  organization={IEEE}
}

@article{zhang2024dd,
  title={DD-Net: Dual decoder network with curriculum learning for full waveform inversion},
  author={Zhang, X. Y. and Min, F. and Pan, S. L. and Xu, Q. and Min, X. Y. and Song, G. J. and Wang, K.},
  journal={IEEE Transactions on Geoscience and Remote Sensing},
  year={2024},
  publisher={IEEE}
}

@article{haghighat2024deeponet,
  title={En-DeepONet: An enrichment approach for enhancing the expressivity of neural operators with applications to seismology},
  author={Haghighat, E. and Bin, W. U. and Karniadakis, G.},
  journal={Computer Methods in Applied Mechanics and Engineering},
  volume={420},
  pages={116681},
  year={2024},
  publisher={Elsevier}
}

@article{rasht2022physics,
  title={Physics-informed neural networks (PINNs) for wave propagation and full waveform inversions},
  author={Rasht-Behesht, M. and Huber, C. and Shukla, K. and Karniadakis, G. E.},
  journal={Journal of Geophysical Research: Solid Earth},
  volume={127},
  number={5},
  pages={e2021JB023120},
  year={2022},
  publisher={Wiley Online Library}
}

@misc{jin2021unsupervised,
  title={Unsupervised learning of full-waveform inversion: Connecting CNN and partial differential equation in a loop},
  author={Jin, P. and Zhang, X. T. and Chen, Y. P. and Huang, S. X. and Liu, Z. C. and Lin, Y. Z.},
  year={2021},
  eprint={2110.07584},
  archivePrefix={arXiv}
}

@article{zhu2022integrating,
  title={Integrating deep neural networks with full-waveform inversion: Reparameterization, regularization, and uncertainty quantification},
  author={Zhu, W. Q. and Xu, K. L. and Darve, E. and Biondi, B. and Beroza, G. C.},
  journal={Geophysics},
  volume={87},
  number={1},
  pages={R93--R109},
  year={2022},
  publisher={Society of Exploration Geophysicists}
}

@article{datta2016estimating,
  title={Estimating a starting model for full-waveform inversion using a global optimization method},
  author={Datta, D.n and Sen, M. K.},
  journal={Geophysics},
  volume={81},
  number={4},
  pages={R211--R223},
  year={2016},
  publisher={Society of Exploration Geophysicists}
}

@article{lin2014acoustic,
  title={Acoustic-and elastic-waveform inversion using a modified total-variation regularization scheme},
  author={Lin, Y. Z. and Huang, L. J.},
  journal={Geophysical Journal International},
  volume={200},
  number={1},
  pages={489--502},
  year={2014},
  publisher={Oxford University Press}
}

@inproceedings{zhang2012wave,
  title={A wave-energy-based precondition approach to full-waveform inversion in the time domain},
  author={Zhang, Z. G and Huang, L. J. and Lin, Y. Z.},
  booktitle={SEG International Exposition and Annual Meeting},
  pages={SEG--2012},
  year={2012},
  organization={SEG}
}

@inproceedings{jin2018learn,
  title={Learn low wavenumber information in FWI via deep inception based convolutional networks},
  author={Jin, Y. C. and Hu, W. Y. and Wu, X. Q. and Chen, J. F.},
  booktitle={SEG International Exposition and Annual Meeting},
  pages={SEG--2018},
  year={2018},
  organization={SEG}
}

@article{sun2021deep,
  title={Deep learning for low-frequency extrapolation of multicomponent data in elastic FWI},
  author={Sun, H. Y. and Demanet, L.},
  journal={IEEE Transactions on Geoscience and Remote Sensing},
  volume={60},
  pages={1--11},
  year={2021},
  publisher={IEEE}
}

@article{yang2023fwigan,
  title={FWIGAN: Full-Waveform Inversion via a Physics-Informed Generative Adversarial Network},
  author={Yang, F. S. and Ma, J. W.},
  journal={Journal of Geophysical Research: Solid Earth},
  volume={128},
  number={4},
  pages={e2022JB025493},
  year={2023},
  publisher={Wiley Online Library}
}

@article{zeng2021inversionnet3d,
  title={InversionNet3D: Efficient and scalable learning for 3-D full-waveform inversion},
  author={Zeng, Q. L. and Feng, S. H. and Wohlberg, B. and Lin, Y. Z.},
  journal={IEEE Transactions on Geoscience and Remote Sensing},
  volume={60},
  pages={1--16},
  year={2021},
  publisher={IEEE}
}

@article{yang2023well,
  title={Well-log information-assisted high-resolution waveform inversion based on deep learning},
  author={Yang, S. L. and Alkhalifah, T. and Ren, Y. X. and Liu, B. and Li, Y. Y. and Jiang, P.},
  journal={IEEE Geoscience and Remote Sensing Letters},
  volume={20},
  pages={1--5},
  year={2023},
  publisher={IEEE}
}

@article{wang2023prior,
  title={A prior regularized full waveform inversion using generative diffusion models},
  author={Wang, F. and Huang, X. Q. and Alkhalifah, T. A.},
  journal={IEEE Transactions on Geoscience and Remote Sensing},
  volume={61},
  pages={1--11},
  year={2023},
  publisher={IEEE}
}

@inproceedings{mardan2024physics,
  title={Physics-Informed Attention-Based Neural Network for Full-Waveform Inversion},
  author={Mardan, A. and Fabien-Ouellet, G.},
  booktitle={85th EAGE Annual Conference \& Exhibition (including the Workshop Programme)},
  volume={2024},
  number={1},
  pages={1--5},
  year={2024},
  organization={European Association of Geoscientists \& Engineers}
}

@article{li2023solving,
  title={Solving seismic wave equations on variable velocity models with Fourier neural operator},
  author={Li, B. and Wang, H. C. and Feng, S. H. and Yang, X. and Lin, Y. Z.},
  journal={IEEE Transactions on Geoscience and Remote Sensing},
  volume={61},
  pages={1--18},
  year={2023},
  publisher={IEEE}
}

@article{lin2023physics,
  title={Physics-guided data-driven seismic inversion: Recent progress and future opportunities in full-waveform inversion},
  author={Lin, Y. Z. and Theiler, J. and Wohlberg, B.},
  journal={IEEE Signal Processing Magazine},
  volume={40},
  number={1},
  pages={115--133},
  year={2023},
  publisher={IEEE}
}

@article{li2023high,
  title={A High Resolution Velocity Inversion Method Based on Attention Convolutional Neural Network.},
  author={Li, W. D. and Liu, H. and Wu, T. Q. and Huo, S. D.},
  journal={IEEE Transactions on Geoscience and Remote Sensing},
  year={2023},
  publisher={IEEE}
}

@article{xu2024aba,
  title={ABA-FWI: Augmented boundary attention for full waveform inversion},
  author={Xu, Q. and Min, F. and Pan, S. L. and Zhang, X. Y. and Song, G. J. and Wang, K. and Wu, X. D.},
  journal={IEEE Transactions on Geoscience and Remote Sensing},
  year={2024},
  publisher={IEEE}
}

@inproceedings{ioffe2015batch,
  title={Batch normalization: Accelerating deep network training by reducing internal covariate shift},
  author={Ioffe, S. and Szegedy, C.},
  booktitle={International conference on machine learning},
  pages={448--456},
  year={2015},
  organization={pmlr}
}

@article{zhang2023seismic,
  title={Seismic inversion based on acoustic wave equations using physics-informed neural network},
  author={Zhang, Y. J. and Zhu, X. Y. and Gao, J. H.},
  journal={IEEE transactions on geoscience and remote sensing},
  volume={61},
  pages={1--11},
  year={2023},
  publisher={IEEE}
}

@inproceedings{finder2024wavelet,
  title={Wavelet convolutions for large receptive fields},
  author={Finder, S. E. and Amoyal, R. and Treister, E. and Freifeld, O.},
  booktitle={European Conference on Computer Vision},
  pages={363--380},
  year={2024},
  organization={Springer}
}

\end{document}